\documentclass{egpubl}
\usepackage{eurovis2024}

\SpecialIssuePaper         

\CGFccby

\usepackage[T1]{fontenc}
\usepackage{dfadobe}  

\usepackage{cite}  
\usepackage{xcolor}
\usepackage{caption}
\usepackage{subcaption}
\usepackage{mathtools}
\usepackage{amsmath}
\usepackage{amssymb}
\usepackage{wrapfig}
\BibtexOrBiblatex
\electronicVersion
\PrintedOrElectronic
\ifpdf \usepackage[pdftex]{graphicx} \pdfcompresslevel=9
\else \usepackage[dvips]{graphicx} \fi

\usepackage{egweblnk}

\title[CUPID]%
{CUPID: Contextual Understanding of Prompt-conditioned Image Distributions}

\author[Y. Zhao \& M. Li \& M. Berger]{
    \parbox{\textwidth}{\centering 
    Y. Zhao\orcid{0000-0002-7214-9619};
    M. Li\orcid{0000-0002-0457-8091}
    and
    M. Berger\orcid{0000-0002-8876-2418} 
    }
    \\
    {\parbox{\textwidth}{\centering 
        Vanderbilt University, USA
    }
    }
}

\DeclarePairedDelimiterX{\infdivx}[2]{(}{)}{%
  #1\;\delimsize\|\;#2%
}
\newcommand{\infdiv}{D_{KL}\infdivx}

\begin{document}


\maketitle

\begin{abstract}
We present CUPID: a visualization method for the contextual understanding of prompt-conditioned image distributions. CUPID targets the visual analysis of distributions produced by modern text-to-image generative models, wherein a user can specify a scene via natural language, and the model generates a set of images, each intended to satisfy the user's description. CUPID is designed to help understand the resulting distribution, using contextual cues to facilitate analysis: objects mentioned in the prompt, novel, synthesized objects not explicitly mentioned, and their potential relationships. Central to CUPID is a novel method for visualizing high-dimensional distributions, wherein contextualized embeddings of objects, those found within images, are mapped to a low-dimensional space via density-based embeddings. We show how such embeddings allows one to discover salient styles of objects within a distribution, as well as identify anomalous, or rare, object styles. Moreover, we introduce conditional density embeddings, whereby conditioning on a given object allows one to compare object dependencies within the distribution. We employ CUPID for analyzing image distributions produced by large-scale diffusion models, where our experimental results offer insights on language misunderstanding from such models and biases in object composition, while also providing an interface for discovery of typical, or rare, synthesized scenes.


\end{abstract}  

\section{Introduction}

Generative AI is becoming commonplace in today's society~\cite{jo2023promise}, due in large part to advancements made in generative models~\cite{saharia2022photorealistic}, and the availability of large-scale datasets~\cite{schuhmann2022laion} on which to train models.
Yet a key reason why generative AI is useful to people is that modern generative models are typically \emph{conditional} in nature.
A conditional generative model synthesizes novel data instances, conditioned on data supplied by a user.
Often such conditioning takes the form of text prompts, e.g. for a text-to-image generative model~\cite{rombach2021highresolution}, a user typically specifies a natural language description of a scene.
The model then conditions on the text prompt in synthesizing a novel image that, ideally, accurately represents the scene envisioned by the user.
The ability to condition on natural language gives users a fluid way to interface with generative models, allowing for these models to be purposed for a variety of creative tasks~\cite{epstein2023art}. For instance, the user might want to produce a single image that meets precise criteria, in contrast with more open-ended pursuits where the user does not know exactly what they want, and thus generative models serve as a source of inspiration.

Independent of downstream task, there remain numerous challenges in deploying text-to-image models for creative purposes.
Specifically, conditioned on a user's prompt, the image generated by the model might differ from what the user has in mind; this has motivated the development of prompt engineering methods~\cite{zhou2022learning,brade2023promptify}.
Moreover, the user may wish to have more fine-grained control on parameters of the generation process, e.g. specifying the importance of objects mentioned in the prompt~\cite{chefer2023attend}, the locations of objects~\cite{xie2023smartbrush}, or object appearance~\cite{epstein2024diffusion}.
Yet once a prompt has been settled on for image generation, and parameters of the generation process fixed, a text-to-image model does not output just a single image.
Rather, these models provide samples drawn from a prompt-conditioned distribution of images, consisting of all possible images that adhere to the user's input(s), subject to being within the training data distribution.
Thus downstream tasks that involve some form of \emph{search} must contend with this distribution, e.g. finding an image satisfying certain criteria, or browsing the distribution to discover general themes and variations of the prompt description.

We argue that a better understanding of the distribution can help facilitate the completion of such tasks.
To this end, we introduce CUPID: a visualization method for the contextual understanding of prompt-conditioned distributions.
CUPID aims to support the understanding of images synthesized by modern text-to-image generative models.
We assume that a user's text prompt describes a scene, comprised of a collection of objects that are intended to appear across generated images.
The primary goal of CUPID is to help users understand the objects that populate the image distribution -- these can be objects specified by the user, or left unspecified, but nevertheless synthesized by the model.
Since the images, and objects therein, are samples drawn from a distribution, we wish to depict the underlying \emph{density} of objects.
A representation of density allows one to understand what is typical, or rare, in a particular object, e.g. the variety in appearance, texture, and style.
Moreover, density representations enable a better understanding of object relationships, e.g. dependencies between objects that are indicative of a potential bias of the model.
CUPID leverages object-specific feature representations~\cite{liu2021swin}, thus allowing us to move beyond image-level language representations~\cite{radford2021learning}, obtaining a finer-grained representation of objects for analysis.

At the heart of CUPID is a method for visually representing the density of objects.
We introduce \emph{density-based embeddings}: a method for producing low-dimensional embeddings that, when visualized, faithfully represents a provided measure of density that has been assigned to the objects.
Given some arbitrary density estimate, for instance the kernel density estimation (KDE)~\cite{parzen1962estimation} of a collection of high-dimensional points, we propose an optimization procedure that ensures the KDE of the low-dimensional embedding is close to that of the provided density estimate.
In turn, a position-based visual encoding of the resulting embedding properly reflects the KDE, e.g. a group of points that are in close proximity corresponds to a region of high density from the original estimate.

A key feature of our embedding method is its generality: we do not make assumptions on how the provided density estimate is computed.
Thus, we can depict the density for a single object, allowing the user to distinguish an object by (1) different styles, and (2) the likelihood of these styles, e.g. typical vs. rare.
But just as importantly, we can compute the joint probability distribution \emph{between} objects, and derive density estimates.
Specifically, we can compute the marginal density of an object, marginalizing over all instances of another object -- this allows us to compare the density of a single object from marginals computed over different objects.
Moreover, we can form conditional densities -- conditioned on an object in a single image, we derive the density for a separate object over all images.
The joint distribution also allows us to compute a measure of dependence between different objects via pointwise mutual information (PMI).
Taken together, these quantities help \emph{contextualize} the analysis: we can study the distribution of a particular type of object within the context of another object, e.g. observing a change in the density, and why such a change might occur.

\begin{figure}[!t]
    \centering   \includegraphics[width=.45\textwidth]{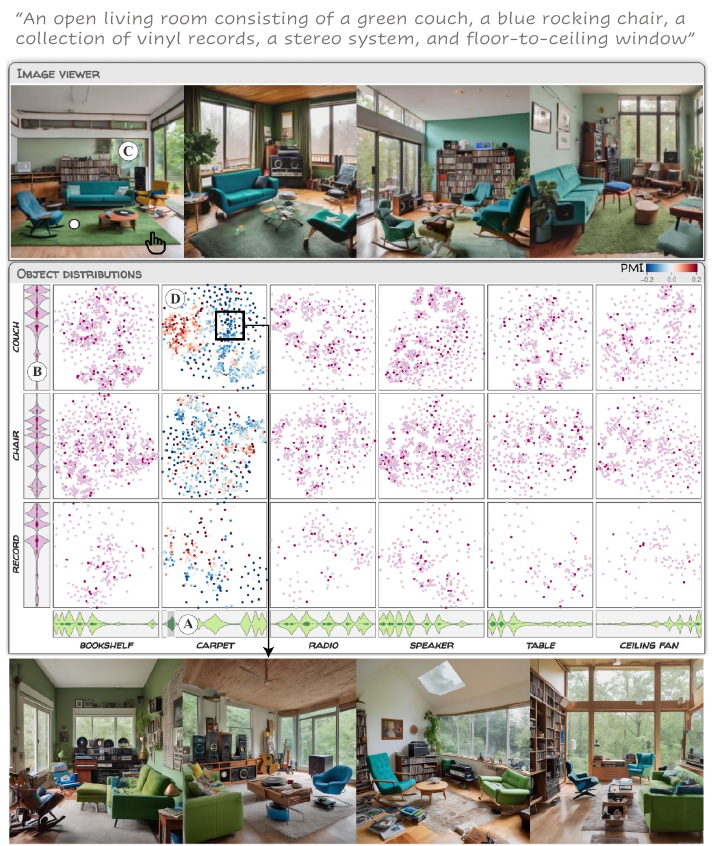}
    \caption{CUPID aims to support users in understanding image distributions produced by generative models. For the given prompt describing the scene (top), one may brush 1D density embeddings of objects unspecified in the prompt, e.g. carpet (A), resulting in linked updates to the remaining views, e.g. 1D densities for specified objects (B). By selecting an image corresponding to the brushed object (C), we can show more detailed information on object relationships (D), here highlighting dependencies that exist between the selected carpet object, and all couches in the distribution.}
    \label{fig:overview}
\end{figure}

CUPID utilizes density-based embeddings within a visual interface for analyzing samples from an image distribution, please see Fig.~\ref{fig:overview}
Objects are organized by those mentioned in the prompt, shown as visual representations of density (left), and objects unspecified, shown on the bottom.
For a prompt object, we encode its marginal density-based 2D embedding as a scatterplot, where we marginalize over each of the unspecified objects, shown as a row of scatterplots.
Linked brushing of a set of objects (A) helps convey the overall distribution across all objects from the provided prompt (record, chair, couch) (B), while for more detailed analysis, selecting a given image of the brushed object (C) enables one to study object relationships.
Specifically, selecting a green carpet in an image results in an update to the 2D density embeddings over all prompt objects (D).
In particular, we visually encode the PMI between each object, and the selected object (green carpet), to reveal dependencies across objects, e.g. PMI with large magnitude indicates dependence.
As one usage of our interface (c.f. Fig.~\ref{fig:overview}), we find that a certain style of one object can \emph{preclude} that same style applied to another object, where the green color specified for couches is instead applied to carpets.

We summarize the contributions of our work:
\begin{itemize}
\item We introduce a technique for computing density-based embeddings. This provides a visual representation of density for objects that exist within a distribution of images.
\item Our technique supports arbitrary density estimates. Thus we can convey the density of individual objects, as well as density estimates derived from the joint distribution of objects, allowing for analysis of object relationships.
\item The embeddings underlie a visualization design for helping understand the image distribution produced by a text-to-image generative model. We demonstrate the utility of our interface in studying object composition, scene perspective, and spatial relationships between objects.
\end{itemize}
\section{Related work}

Our work spans several areas in visualization and HCI: using generative models to create images, analyzing such models, methods for exploring images, as well as high-dimensional data analysis methods. We discuss each in turn.

\subsection{Human-AI co-creation of images}

The rise of text-to-image generative models~\cite{rombach2021highresolution,ramesh2022hierarchical,saharia2022photorealistic} has sparked numerous interface designs~\cite{son2023genquery,brade2023promptify,feng2023promptmagician,suh2023structured} to help humans co-create with generative models -- even for the creation of visualizations~\cite{Schetinger2023DoomDeliciousnessChallenges,wu2023viz2viz}. Some designs focus on editing a single image at a time~\cite{zhang2023adding,chung2023promptpaint} using methods that either control the sampling process of a diffusion model, or explicitly train a new diffusion model for a target mode of interaction. Other designs are more exploratory, and study different ways of querying for images~\cite{son2023genquery}, exploring the overall design space that might be offered by a generative model~\cite{suh2023structured}, as well as visual interfaces for exploring a collection of prompts in the pursuit of finding a desired image~\cite{brade2023promptify,feng2023promptmagician}. In particular, Promptify~\cite{brade2023promptify} and PromptMagician~\cite{feng2023promptmagician} are closely related to our work, in enabling users to fluidly explore images produced by text-to-image models. However, such methods use either (1) image-level features, or (2) text-to-image alignment scores, e.g. CLIP~\cite{radford2021learning}. Our work goes beyond image-level representations, and instead relies on object-level representations to offer a more detailed view on the distribution of images produced by a generative model. In this sense, we view CUPID as complementary to these works, along with other like-minded methods that rely on navigating a large collection of images.

Another important distinction in our work is the study of a \emph{single} prompt-conditioned image distribution. The incentive to explore \emph{multiple} prompts is due to the unpredictability of text-to-image generative models, given a prompt describing a scene. Prior work~\cite{zamfirescu2023johnny} has studied how prompting, more broadly for generative models, can be challenging to adequately perform. Thus, numerous methods for prompt engineering~\cite{pryzant2023automatic,hao2022optimizing} exist to translate a user's intent to a prompt suitable for conditioning on a generative model. On the other hand, often, prompts are not intended to be highly precise descriptions that yield a single unique image. And so understanding what a single prompt-conditioned distribution has to offer can complement methods that further seek to refine a prompt for model alignment.

\subsection{Analyzing text-to-image generative models}

Beyond the use of CUPID for search or discovery tasks, our interface is also suitable to help verify whether the image distribution is faithful to the provided prompt. This process of verification is related to existing work that focuses on analyzing the capabilities of text-to-image generative models. In particular, numerous methods have demonstrated that text-to-image diffusion models often misunderstand the intent behind a user's prompt, e.g. to synthesize and compose objects~\cite{patel2023conceptbed,okawa2023compositional}, adhere to spatial reasoning~\cite{gokhale2022benchmarking}, and reason about relationships amongst objects~\cite{conwell2022testing}. To a certain extent, it is possible to use existing models in measuring the adherence of a synthesized image to a given prompt~\cite{yarom2023you,karthik2023if} , and there is evidence that the learned representations of vision-language models carry meaningful semantics~\cite{goh2021multimodal}). Nevertheless, vision-language reasoning still remains challenging~\cite{ma2023crepe}, while such existing benchmarks usually report coarse summaries on alignment~\cite{ma2023crepe} or composition abilities~\cite{patel2023conceptbed}. In contrast, CUPID offers an interactive means of studying potential failure modes, or biases, that might exist in an image distribution, without relying on predefined textual probes to discover such issues. In particular, certain features of objects can be difficult to describe with language to begin with, e.g. the type of shape of an object, and thus our use of object-based feature representations~\cite{liu2021swin} permits a more fine-grained analysis of image distributions.

\subsection{Exploring image distributions}

The visualization community has extensively studied methods for exploring collections of images, and visual data more broadly -- please see~\cite{afzal2023visualization} for a survey. Closely related to CUPID are methods for leveraging both images and text for exploring image collections~\cite{gu2015igraph,xie2018semantic}. However, these methods utilize image-level features, and thus for more complex scenes they might fail to capture the diversity of objects, and their properties. Numerous techniques exist for more flexibly exploring images, e.g. PhotoMesa~\cite{bederson2001photomesa} for more expressive zoom, as well as methods that target specific kinds of analysis in exploration, e.g. interactively refining a recommendation system~\cite{zahalka2020ii}, or the study of image datasets as they pertain to training machine learning models~\cite{9904448}. These methods can offer a deeper analysis on image collections; nevertheless, our method is distinct from prior works in supporting exploration at the object-level, rather than focusing on image-level exploration.

\subsection{High-dimensional visual analysis}

A central problem in our work is how to explore samples drawn from a high-dimensional probability distribution, and in particular, different perspectives of the distribution, realized as high-dimensional feature vectors that correspond to specific objects. In many visual analytics systems, a combination of dimensionality reduction (DR) and clustering is often used~\cite{wenskovitch2017towards,xia2022interactive} to support potential patterns in the data (e.g. tight cluster, or intra-cluster similarity). In our scenario, as data are drawn from a probability distribution, our analysis is driven by estimating, and visualizing, the density~\cite{van2017variable,backurs2019space} of objects in individual scenes (images). In this setting, we make fewer assumptions about our data compared to clustering~\cite{tran2006knn}, while still prioritizing the density estimates we compute when deriving a low-dimensional embedding of objects, akin to DR methods. In particular, DR techniques have considered how to utilize similar forms of side information, e.g. contrastive DR~\cite{fujiwara2019supporting} or class-constrained t-SNE~\cite{meng2023class}. In particular, our approach is closely related to Meng et al., but rather than consider class probabilities in deriving embeddings, we use a more general form of information -- density estimates -- and a method for ensuring that the low-dimensional embedding preserves density. This is in contrast with the more traditional use of density estimation, as applied to a 2D projection~\cite{pezzotti2016approximated}, to better emphasize clusters inferred by a DR method.
\section{CUPID objectives and data}
\label{sec:design}

In this section we discuss our problem setting and the objectives of CUPID, and provide details on the data that we collect for visualization.

\begin{figure}[!t]
    \centering
    \includegraphics[width=.35\textwidth]{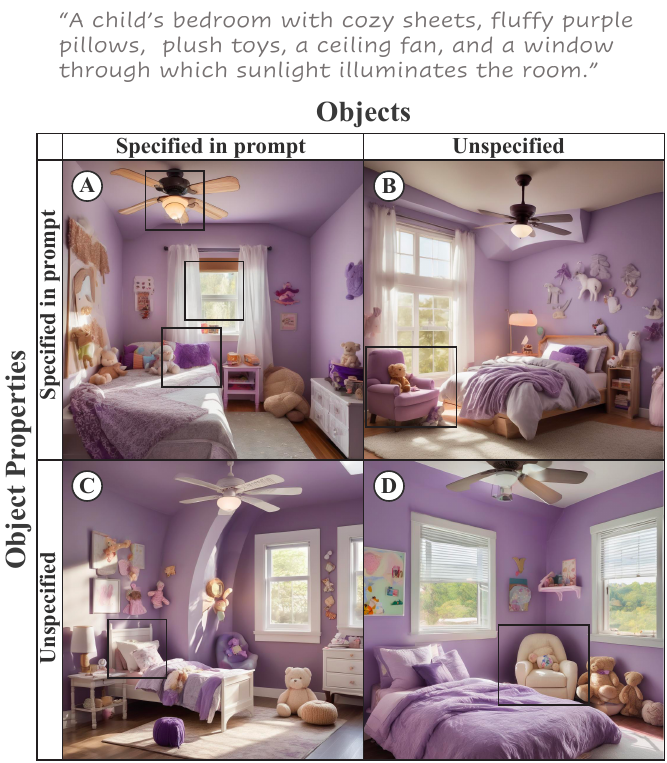}
    \caption{Our analysis of prompt-conditioned image distributions is organized around (1) objects that appear in the distribution, whether specified in the prompt or not, and (2) the properties of objects, specified or unspecified. Ideally, generated images are consistent with the prompt (A), while still exhibiting diversity characteristic of the scene (D). Issues that arise can be due to language misunderstanding, whether omitting a specified object's property (C), or biasing the properties of an unspecified object (B).}
    \label{fig:setup}
\end{figure}

We are primarily interested in studying what is produced by generative text-to-image models.
In particular, CUPID is motivated by modern text-to-image diffusion models~\cite{rombach2021highresolution} that are trained on large-scale image-captioned datasets, e.g. of the order of billions of images~\cite{schuhmann2022laion}.
A text-to-image diffusion model is a \emph{conditional} probability distribution over images, where for some text prompt denoted $\mathbf{c}$, a diffusion model provides a means of drawing images from a conditional distribution: $\mathbf{x} \sim p_{\theta}(\mathbf{x} | \mathbf{c})$, where $\mathbf{x}$ is an image, and $\theta$ denotes model parameters.
Samples are drawn from this distribution by using a deep UNet-based convolutional network~\cite{rombach2021highresolution} to transform a presented noise image into a realistic-looking image.
Thus, a sampling of noise images -- each often associated with an integer used to seed a pseudo-random number generator -- gives rise to samples from the conditional distribution.
Assuming the model is of sufficiently high capacity, the prompt-conditioned distribution $p_{\theta}(\mathbf{x} | \mathbf{c})$ will assign high density to images that are (1) faithfully described by the prompt $\mathbf{c}$, and (2) representative of the training data distribution~\cite{kingma2021variational}.

To structure our analysis, we make a few assumptions on the types of prompts that are provided, and what we expect to see in the resulting distribution.
We assume the prompt $\mathbf{c}$ corresponds to some arbitrary scene, e.g. an outdoor environment or an indoor setting such as a bedroom.
We organize a scene along two axes: the \textbf{objects} within a scene, and the \textbf{properties} of objects, e.g. this could be their appearance, size, pose, texture, etc..
Moreover, the objects within a scene either correspond to objects specified in the prompt, or \emph{unspecified} objects not mentioned in the prompt, but nevertheless synthesized by the generative model.
Likewise, the properties of an object may, or may not, be consistent with how the object is described in the prompt.
Last, unspecified objects might have arbitrary properties, or alternatively take on properties that are described in the prompt, but made in reference to other objects.
We summarize these scenarios in Fig.~\ref{fig:setup}.

Given the uncertainty in what a generative model will produce when provided a prompt, CUPID targets two types of analyses: verification, and discovery.
\begin{itemize}
    \item \textbf{Verification:} at a basic level, is the generative model producing images that are consistent with the given prompt? A user might wish to verify the distribution's faithfulness to the prompt's objects, along with their properties (c.f. Fig.~\ref{fig:setup}(A)). A mismatch can occur when an object only appears in a subset of images. Moreover, when an object does appear, its properties may differ from what was specified (c.f. Fig.~\ref{fig:setup}(C)).
    \item \textbf{Discovery:} for objects whose properties are vague, or left unstated, a user may want to understand what properties are synthesized by the generative model, e.g. what did the model decide on, in the absence of the user deciding on object properties. Moreover, the model can generate unspecified objects (c.f. Fig.~\ref{fig:setup}(D)). Ideally these objects are characteristic of the scene, and their properties are not biased by the properties specified in the prompt (c.f. Fig.~\ref{fig:setup}(B)).
\end{itemize}

To support these analyses, CUPID seeks to obtain a deeper understanding of the distribution $p_{\theta}(\mathbf{x} | \mathbf{c})$, moving beyond just the density of an image, and instead, estimating the density of an object within an image. Object-based density can help verify what is typical or rare in the properties of an object. Moreover, the joint distribution over multiple objects can help us understand dependencies between objects. This can be used to better understand how objects compose with one another, allowing for a more fine-grained organization of object properties, while also indicating potential biases in the composition of a scene. The notion of a ``property'' of an object, however, may not be easily expressed in language, and thus for density estimation, CUPID leverages object-level feature representations extracted in images.

In detail, CUPID uses the text-to-image diffusion model of SDXL~\cite{podell2023sdxl}, which produces images of resolution $1024 \times 1024$.
We use this model to ensure synthesized objects in images are of sufficient resolution.
Given a prompt, we generate a collection of images from SDXL $\mathbf{x}_n \sim p(\mathbf{x} | \mathbf{c})$ for a total of $N$ images, where $n \in [1,\ldots,N]$.
From this set of images, our aim is to find all possible objects that have been synthesized by the model, across all images.
To do so, we use RAM~\cite{zhang2023recognize}, or the recognize anything model, for open-vocabulary object recognition.
Provided an arbitrary set of object labels, for a given image, RAM predicts a set of labels that correspond to objects that exist in the scene.
The labels are post-processed, in order to distinguish objects in the prompt from those unspecified, while labels that carry similar semantics are merged, in order to obtain a more concise label set -- we defer these details to supplementary.

The mere presence of an object within an image, however, is insufficient for us in understanding the specific properties that can object takes on.
To address this, we derive an object-based representation for every object recognized in an image, building on the RAM model.
Specifically, RAM uses multi-headed cross-attention~\cite{vaswani2017attention} between spatially-referenced image features and a label's text features when predicting whether a given label is present in an image.
We use this cross-attention as a way to \emph{localize} where in the image an object has been recognized.
Given that the cross-attention used by RAM is relatively lightweight -- only 2 layers, with 4 attention heads -- we find that averaging the last layer's cross-attention maps, over all attention heads, gives an effective way to localize an object.
Specifically, we find the position in the image feature map whose averaged cross-attention is largest, and associate the labeled object with this location.
We then take the visual feature at this location~\cite{liu2021swin} -- what is provided as input to cross-attention -- as the object-based representation.

In summary, for each image $\mathbf{x}_n$, we assume a set of objects $\mathcal{S}_n$ has been detected via RAM, and we associate each object $s \in \mathcal{S}_n$ with a feature vector denoted $\mathbf{z}_{n,s}$.
The full collection of objects is provided via $\mathcal{S} = \bigcup_{n=1}^N \mathcal{S}_n$, which we distinguish between (1) objects specified in the prompt, and (2) objects unspecified.
Importantly, an object is not represented by language alone, but rather, through feature representations of a pretrained vision transformer model~\cite{liu2021swin}.
\section{Density-based embeddings}

\begin{figure}[!t]
    \centering
    \includegraphics[width=.4\textwidth]{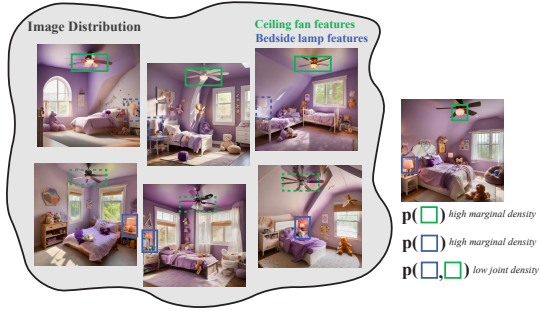}
    \caption{Density estimation enables us to quantify what is typical, and what is rare, within a given image distribution. For instance, ceiling fan objects whose lights are turned on can have similar feature representations, and thus report high density. In contrast, ceiling fans and lamps whose lights are both turned on will report a low joint density, thus reflecting a rare object relationship.}
    \label{fig:density_illustration}
\end{figure}

In this section we describe the methods used that underlie CUPID's visualization.
As the mathematical object under study is a (conditional) probability distribution, our aim is to estimate the density of samples that have been drawn from this distribution, and subsequently, visualize the density estimate.

\subsection{Object-based density estimation}

Numerous likelihood-based models exist for either exactly~\cite{papamakarios2021normalizing} or approximately~\cite{kingma2013auto} reporting the density of high-dimensional data, yet such methods are aimed at generalization.
In contrast, we are only concerned with a collection of images that have been generated by a diffusion model, and obtaining density estimates restricted to the provided set.
To this end, we propose to use nonparametric density estimation methods~\cite{parzen1962estimation}, applied to individual objects synthesized by a generative model.
Specifically, we use KDE to obtain an unnormalized probability distribution over objects via:
\begin{equation}
    \tilde{P}_d(\mathbf{z}_{i,s}) = \sum_{n \in O_s} k\Big(\frac{\lVert \mathbf{z}_{i,s} - \mathbf{z}_{n,s} \rVert^2}{h} \Big),
\label{eq:kde}
\end{equation}
where $s \in \mathcal{S}$ represents an object, $i \in [1,\ldots,N]$ indexes an image from the distribution, and $O_s$ is the set of images where object $s$ was detected.
The function $k$ is the kernel, with associated bandwidth $h$.
Intuitively, $h$ controls for the distance at which a pair of vectors $\mathbf{z}_{i,s}$ and $\mathbf{z}_{n,s}$ are considered similar.
We take $k$ to be an exponential kernel, e.g. $k(z) = \exp[-z]$ for input $z > 0$.

We can further extend density estimation of single objects, to multiple objects, through forming a joint distribution. This is useful for modeling object relationships. In particular, we define the joint KDE to be:
\begin{equation}
    \tilde{P}_d(\mathbf{z}_{i,s},\mathbf{z}_{j,t}) = \sum_{n \in O_s \cap O_t} k\Big(\frac{\lVert \mathbf{z}_{i,s} - \mathbf{z}_{n,s} \rVert^2}{h} \Big) \cdot k\Big(\frac{\lVert \mathbf{z}_{j,t} - \mathbf{z}_{n,t} \rVert^2}{h} \Big),
\end{equation}
where $s,t \in \mathcal{S}$ correspond to distinct objects, while $i$ and $j$ index over images. Within the sum, we only report a high count for image $n$ shared by the objects ($O_s \cap O_t$) if, both, object $s$ for image $i$ is close to the corresponding object in image $n$, and object $t$ for image $j$ is close to its corresponding object in image $n$.

From the unnormalized distributions, we may easily normalize each, through dividing by their sum of values.
This gives us probability distributions over individual objects $P_d(\mathbf{z}_{i,s})$ and joint probability distributions over pairs of objects $P_d(\mathbf{z}_{i,s},\mathbf{z}_{j,t})$.
The unnormalized distributions, however, have the interpretation of performing ``soft'' counts over images, e.g. for $\tilde{P}_d(\mathbf{z}_{i,s})$ this would be the number of times an object $s$ with representation $\mathbf{z}_{i,s}$ occurs in the distribution.
This enables us to compare the various properties of an object through their density estimates: objects that have typical properties (e.g. common appearance, texture) will give high counts, while objects that are rare in the distribution will report low counts.
Fig.~\ref{fig:density_illustration} highlights the scenario in which we are considering the joint probability distribution $P_d(\mathbf{z}_{i,s},\mathbf{z}_{j,t})$.
Here, marginals will be reported as high probability for the lamps, and ceiling fans, in both having lights turned on, relative to the full image distribution.
However, \emph{jointly}, objects with such properties rarely co-occur, and thus their joint distribution will report a low probability.

\subsection{Density-preserving low-dimensional object embeddings}

Given a probability distribution that characterizes an object indexed by $s$, denoted $P_d(\mathbf{z}_{i,s})$, we would like to visualize the distribution.
The main purpose is to convey the modes of the distribution, e.g. what is typical, and what is rare.
Though visually conveying a density-based clustering~\cite{tran2006knn} is one such option, it is not clear if well-delineated clusters exist in the distribution to begin with.
Dimensionality reduction (DR) would be a less-restrictive way to show the data, but might not accurately capture the density we have computed.
Instead, we propose a method that strikes a balance between clustering and DR: we aim to derive a low-dimensional embedding for each image associated with an object, such that the (normalized) KDE of the low-dimensional embedding matches the provided density estimate.

To this end, we aim to find an embedding denoted $\mathbf{X}_s \in \mathbb{R}^{|O_s| \times D}$ for object $s$, and embedding dimensionality $D \in \{1,2\}$.
To do so, we minimize the KL divergence between normalized density estimates:
\begin{equation}
    \min_{\mathbf{X}_s} \infdiv{P_d}{Q_d} = \min_{\mathbf{X}_s} \sum_{n \in O_s} P_d(\mathbf{z}_{n,s}) \log \frac{P_d(\mathbf{z}_{n,s})}{Q_d(\mathbf{x}_{n,s})},
    \label{eq:kl-density}
\end{equation}
where $Q_d(\mathbf{x}_{n,s})$ is the normalized version of the KDE, defined in the same way as Eq.~\ref{eq:kde}, replacing the object features with their low-dimensional counterparts.
We set $h=1$ for $Q_d$, as the bandwidth merely fixes the scale of the embedding space.

An issue with minimizing Eq.~\ref{eq:kl-density} is that it can be underconstrained, e.g. there are many potential embedding configurations that would lead to similar KDE-based probability distributions.
Thus, we combine this scheme with a neighbor embedding method, namely tSNE~\cite{van2008visualizing}, to ensure that local neighborhoods of points in the high-dimensional space are preserved in the low-dimensional space.
Specifically, given a neighbor probability distribution over pairs of high-dimensional data points $P_n$, and a corresponding distribution governing the low-dimensional embedding $Q_n$, we solve for the following objective:
\begin{equation}
    \min_{\mathbf{X}_s} \infdiv{P_d}{Q_d} + \lambda \infdiv{P_n}{Q_n},
    \label{eq:kl-both}
\end{equation}
where $\lambda$ balances the contributions of the objectives.
We set $\lambda = 0.1$ for all experiments, thus giving priority to the density criterion.

\begin{figure}[t]
     \centering
     \begin{subfigure}[h]{0.21\textwidth}
         \centering
         \includegraphics[width=.8\textwidth]{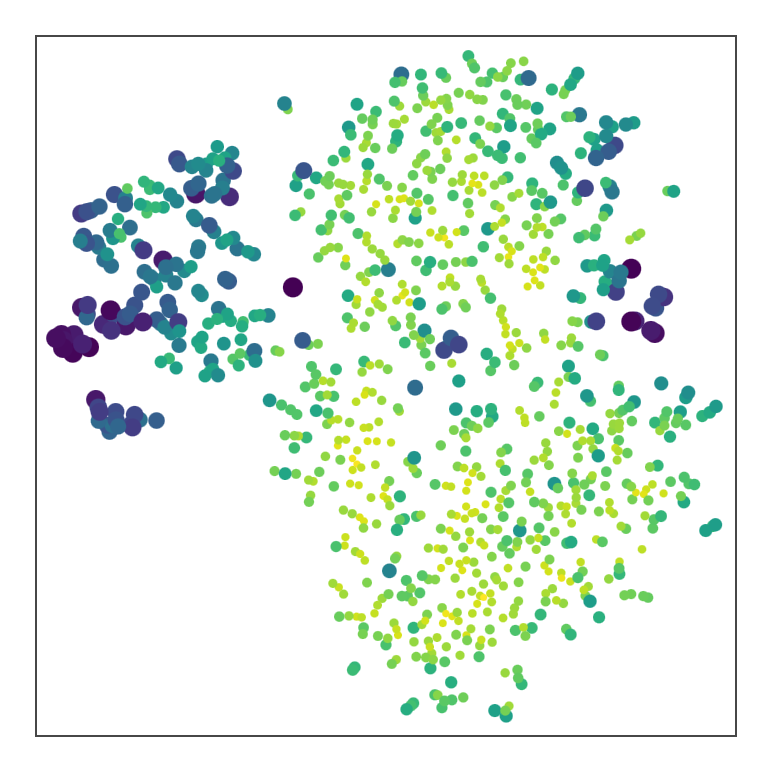}
         \caption{tSNE: $\infdiv{P_d}{Q_d} = .18$}
         \label{fig:tsne_35}
     \end{subfigure}
     \begin{subfigure}[h]{0.21\textwidth}
         \centering
         \includegraphics[width=.8\textwidth]{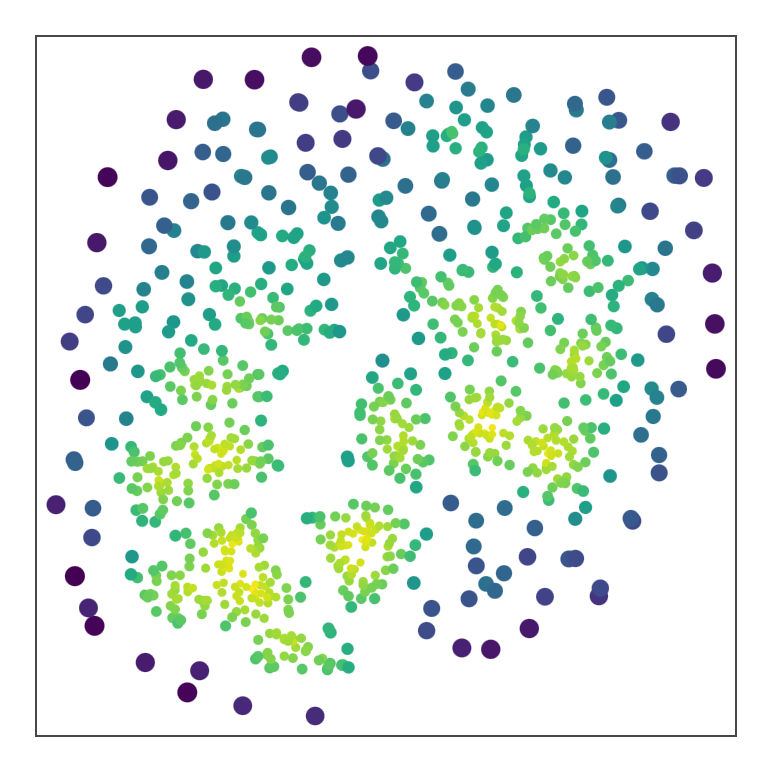}
         \caption{dSNE: $\infdiv{P_d}{Q_d} = .004$}
         \label{fig:density_35}
     \end{subfigure}
     \begin{subfigure}[h]{0.21\textwidth}
         \centering
         \includegraphics[width=.8\textwidth]{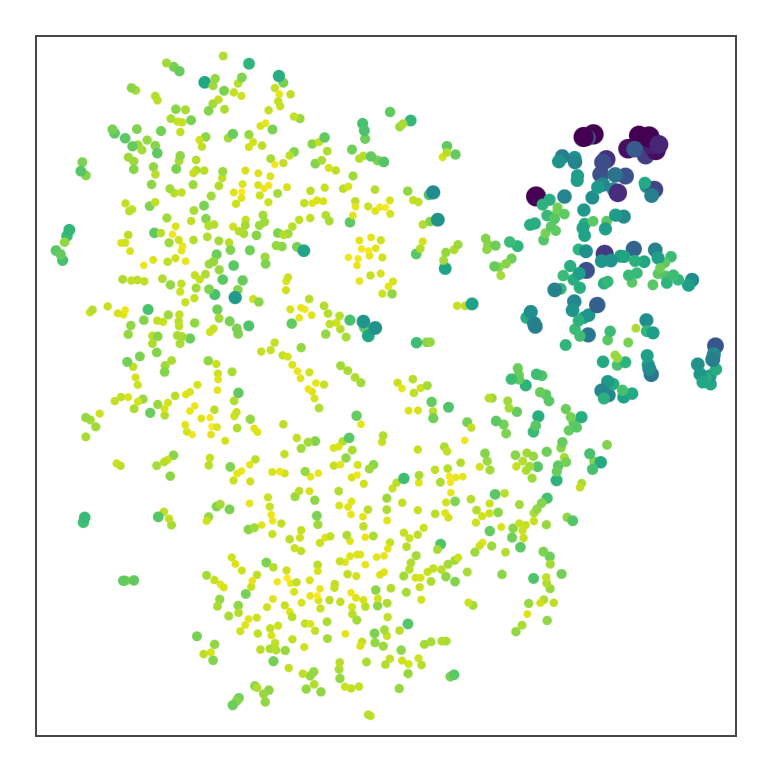}
         \caption{tSNE: $\infdiv{P_d}{Q_d} = .07$}
         \label{fig:tsne_85}
     \end{subfigure}
     \begin{subfigure}[h]{0.21\textwidth}
         \centering
         \includegraphics[width=.8\textwidth]{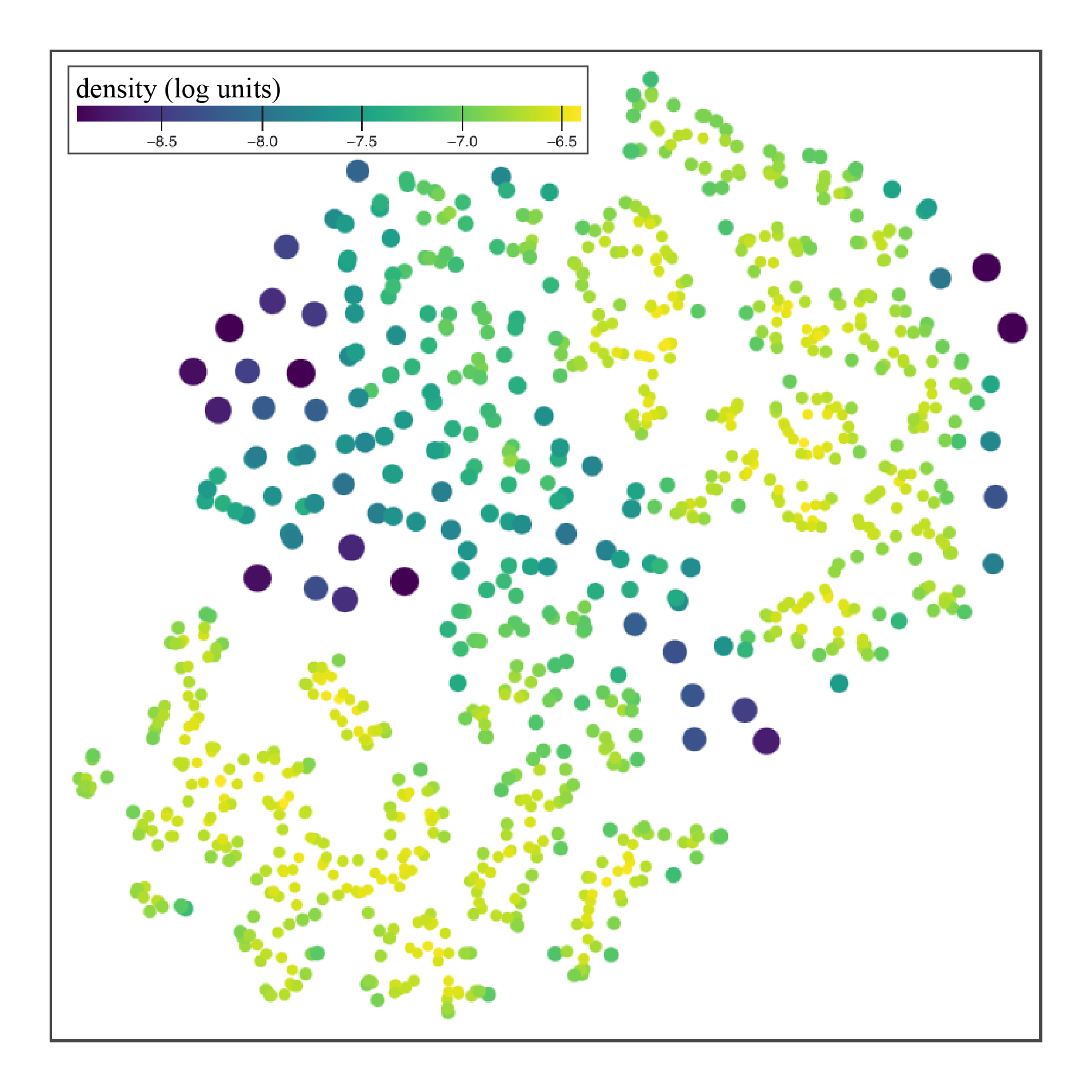}
         \caption{dSNE: $\infdiv{P_d}{Q_d} = .002$}
         \label{fig:density_85}
     \end{subfigure}
        \caption{We compare our density-based embedding approach (dSNE) to that of tSNE, where the provided density is both size and color-encoded in the plots. Across varying bandwidths in the KDE ($h = 40$ top row, $h= 80$ bottom row), our method obtains superior results in density preservation.}
        \label{fig:density-tsne}
\end{figure}

One may ask whether tSNE alone is sufficient in preserving the density of the original data.
To test this, we compare our method -- what we term dSNE for brevity -- to tSNE, with varying bandwidth $h \in \{40,80\}$, as applied to ``chair'' features in the prompt-conditioned distribution within Fig.~\ref{fig:overview}.
We show the results in Fig.~\ref{fig:density-tsne}, where points are both colored, and sized, based on their (log) probabilities of the target distribution $P_d$.
As reported in the KL divergence values (c.f. Eq.~\ref{eq:kl-density}), and the clearer relationship between point density in the scatterplots and the provided density values, our method is able to faithfully capture the original density.
In these examples, both tSNE and dSNE compute neighborhood bandwidths based on a prescribed perplexity~\cite{van2008visualizing}, which might give bias towards factoring out local density variations.
However, in fixing the neighborhood bandwidths to a constant $h$ for tSNE, namely the same value used for KDE, we in fact found worse results, indicating that neighbor embedding methods are inappropriate for capturing density.

A key feature of our embedding method is that it does not depend on how the provided density estimate was computed.
We only assume a normalized probability distribution, defined over the instances of an object, as input.
This allows for flexibility in the types of density estimates we can visually represent.

\begin{figure*}[t]
\centering
\begin{subfigure}[b]{0.3\textwidth}
\includegraphics[width=\textwidth]{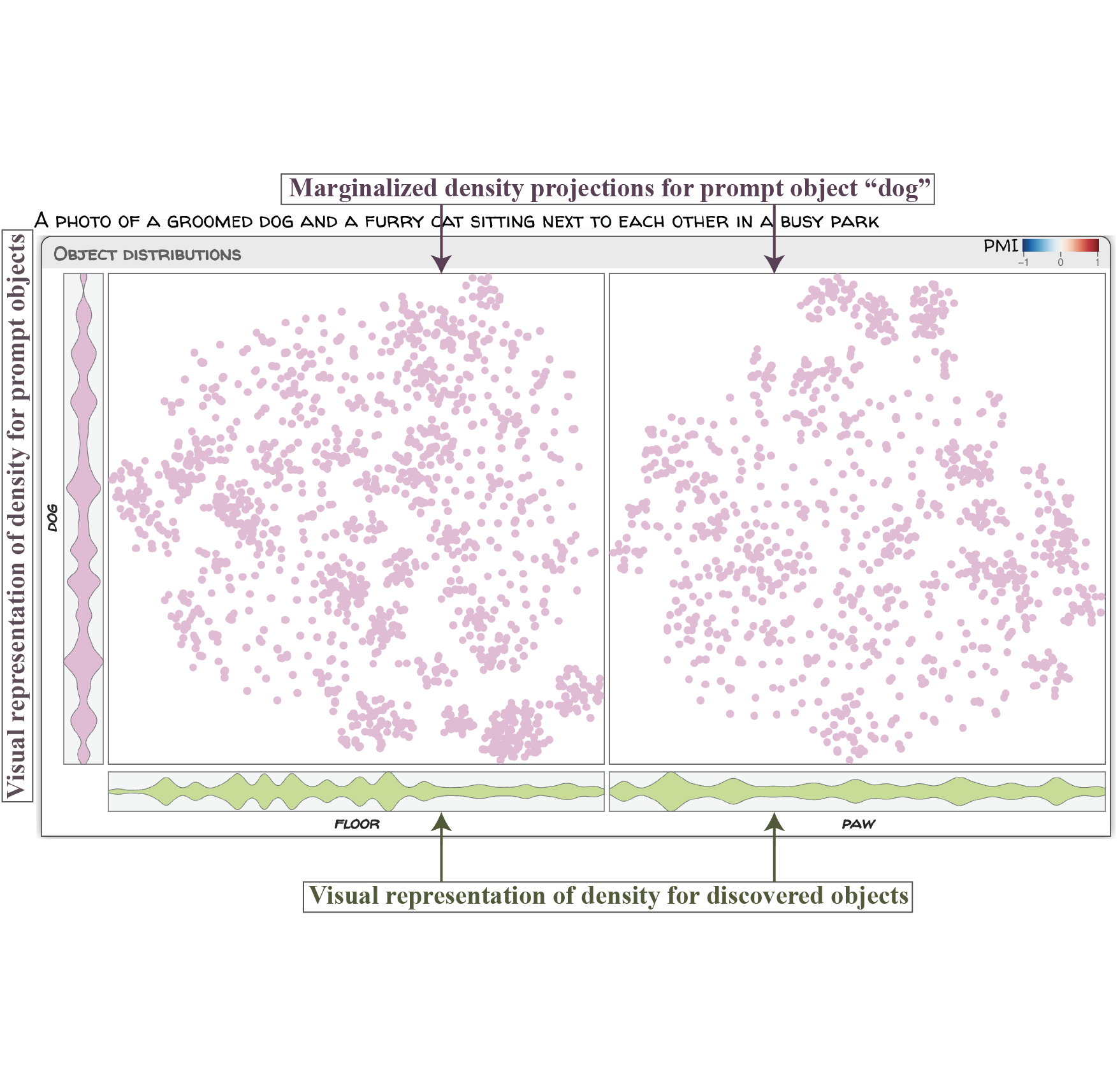}
\caption{Visualization design overview.}
\label{fig:design-a}
\end{subfigure}
\begin{subfigure}[b]{0.3\textwidth}
\includegraphics[width=\textwidth]{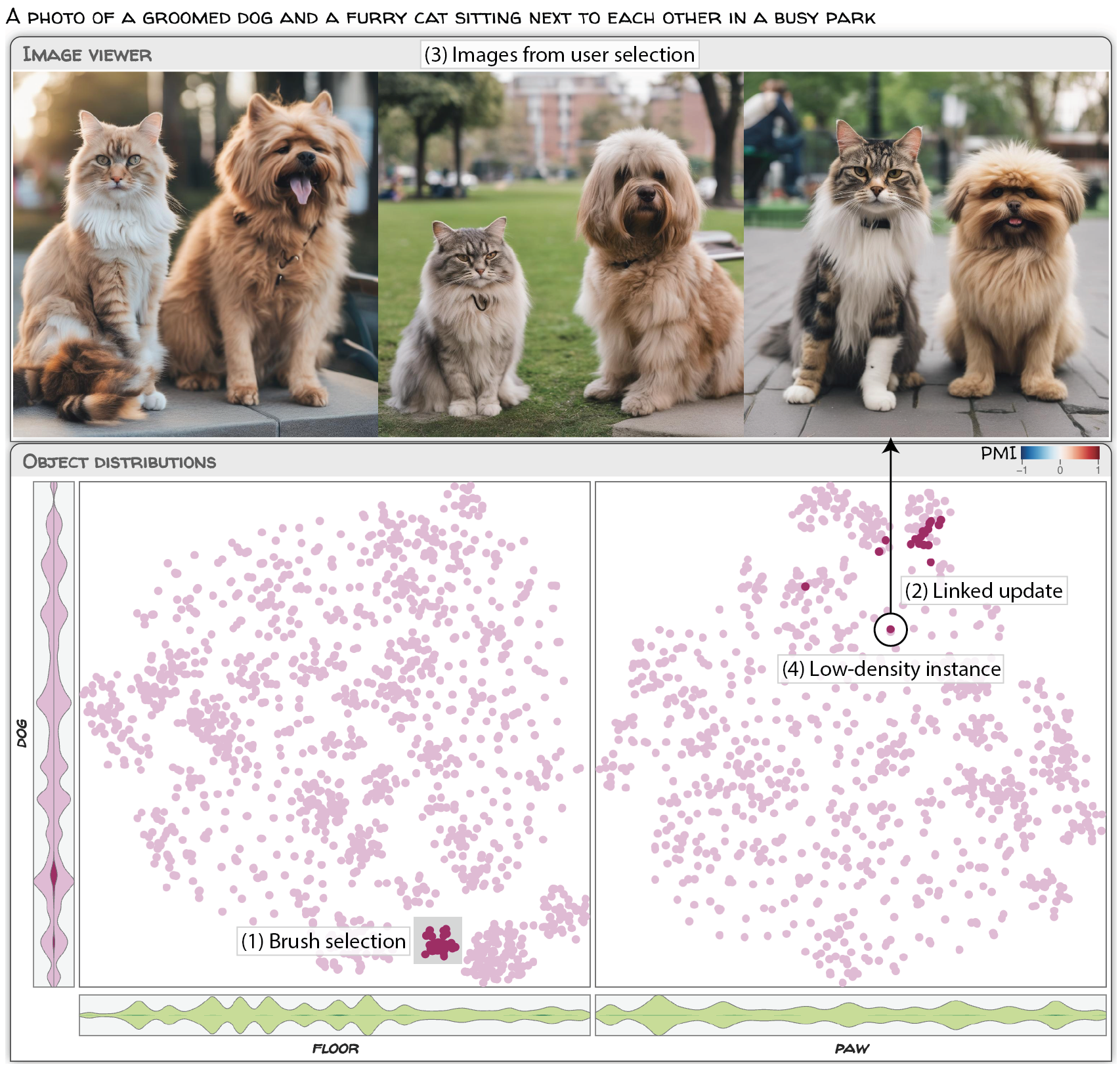}
\caption{Marginal density interaction.}
\label{fig:design-b}
\end{subfigure}
\begin{subfigure}[b]{0.3\textwidth}
\includegraphics[width=\textwidth]{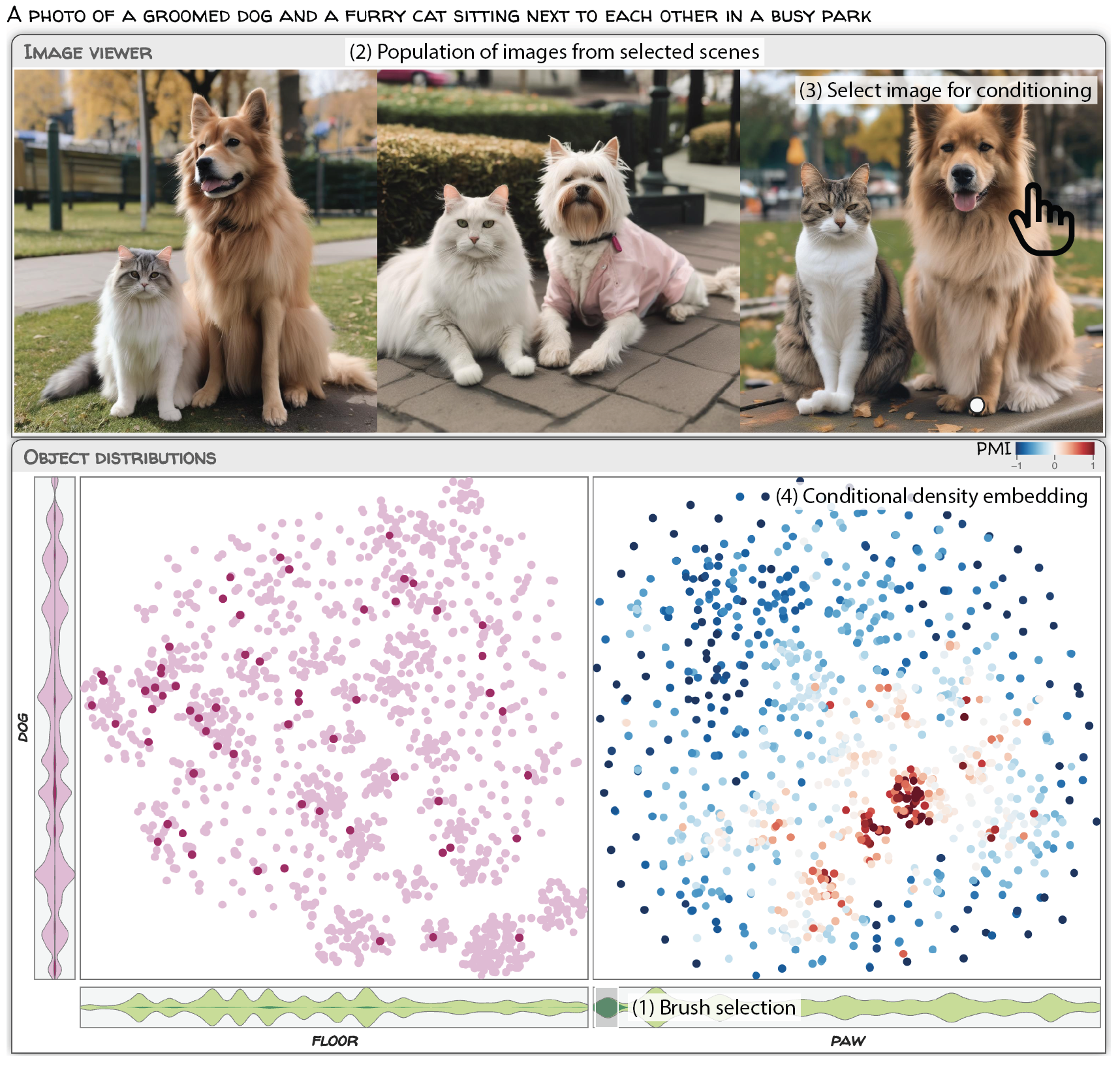}
\caption{Conditional density interaction.}
\label{fig:design-c}
\end{subfigure}
\caption{An overview of the design for CUPID. (a) Density-based object embeddings are shown as violin plots for both prompt objects and discovered objects, with object-marginalized density embeddings shown as 2D scatterplots. (b) Our interface supports linked brushing to relate density representations, in this instance highlighting a rare Shih Tzu that has an angular limb deformity. (c) Further, the interface allows for projection steering via conditioning on a selected image of a particular object, here for the paw feature of the chosen dog.}
\label{fig:design}
\end{figure*}

\textbf{Individual object densities:} this reflects the most straightforward scenario, namely the density $P_d(\mathbf{z}_{i,s})$ is the normalized KDE of the object's collection of feature vectors, in proportion to Eq.~\ref{eq:kde}. An embedding of this density provides a view exclusive to this object, namely, a grouping of the object's properties, and the likelihood of a group appearing within the sampled distribution. Hence, this form of density can help convey the more common properties that an object takes on, as well as its more rare properties.

\textbf{Marginalized object densities:} the provided densities need not be limited to individual objects. Instead, we can derive densities based on the joint distribution of pairs of objects, in order to help reveal object relationships. In particular, for an object indexed by $s$, we can \emph{marginalize} out all instances of another object, indexed by $t$, giving us the following marginal densities:
\begin{equation}
    P_d(\mathbf{z}_{i,s} ; t) = \sum_{n \in O_s \cap O_t} P_d(\mathbf{z}_{i,s}, \mathbf{z}_{n,t}),
\label{eq:marginal}
\end{equation}
where the notation $(; t)$ indicates a dependency on object $t$.
The depiction of marginal densities is most useful when we wish to better distinguish object properties, helping identify object properties that are common when considering one object ($t$), but rare when considering a different object ($t'$).
Namely, marginalizing out an object indexed by $t$ can reveal a high-density grouping in the resulting embedding, but this very group might present low densities when marginalizing over a different object $t'$.

\textbf{Conditional densities:} rather than marginalizing over all instances of an object, we can allow for the user to select an instance, indexed by $j$, for a particular object, indexed by $t$. We may then form a conditional density, conditioned on this particular object's instance:
\begin{equation}
    P_d(\mathbf{z}_{i,s} | \mathbf{z}_{j,t}) = \frac{P(\mathbf{z}_{i,s} , \mathbf{z}_{j,t})}{P_d(\mathbf{z}_{i,s} ; t)}.
\label{eq:marginal}
\end{equation}
This leads to a means of \emph{steering} projections based on user-selected objects.
Intuitively, if we find the conditional density leads to an embedding that does not differ too much from the marginal, then this is an indication of independence between objects.
Otherwise this may indicate a level of dependence on the chosen object instance, e.g. the density might be more spread out, or alternatively, new regions of high density might form.

Aside from finding embeddings, we can also compute and communicate measures of (in-)dependence regarding objects.
Specifically, we may compute the pointwise mutual information (PMI) between objects:
\begin{equation}
PMI(\mathbf{z}_{i,s};\mathbf{z}_{j,t}) = \log P_d(\mathbf{z}_{i,s}, \mathbf{z}_{j,t}) - (\log P_d(\mathbf{z}_{i,s}) + \log P_d(\mathbf{z}_{j,t})).
\end{equation}
If the PMI is zero, this suggests that the objects are independent of one another.
Otherwise, a PMI of high magnitude can elucidate potential biases in how scenes are composed.
Namely, a positive PMI indicates two different kinds of objects that frequently co-occur throughout the distribution.
Conversely, negative PMI indicates objects that rarely co-occur.
\section{CUPID visualization design}

We utilize the density-based visualization techniques as part of a visualization design for exploring image distributions.
The design is motivated by (1) finding different properties of objects across the distribution, for objects specified in the prompt as well as those unspecified, or what we term \textbf{discovered objects}, and (2) finding relationships between objects -- please see Fig.~\ref{fig:design} for an illustration.

\textbf{Object density encodings:} CUPID finds 1-dimensional density-based embeddings for each object, in order to derive object-specific positional encodings of scenes.
We opt to use violin plots to show the density, wherein a Gaussian KDE, with bandwidth $h=1$ is performed on the 1D embeddings, and the resulting density encoded as width in an area mark.
Importantly, this \emph{directly} matches what we optimize for in our density-based embeddings, and thus is a faithful visual encoding of the density.
We find these embeddings using the 1D probability distributions, rather than joint distributions, to give a summary over all scenes.
This process is performed individually for objects that have been discovered by RAM -- to distinguish from prompt objects, we color the area marks green, and then horizontally position the violin plots.
The same process is performed for prompt objects as well, wherein we vertically arrange the violin plots, please see Fig.~\ref{fig:design-a}.

\textbf{Marginalized density encodings:} for every combination of prompt object and discovered object, we display a matrix of 2D scatterplots, with each view depicting the \emph{marginalized density}. Specifically, rows correspond to objects mentioned in the prompt, while columns correspond to discovered objects. A single view in this matrix corresponds to a 2D density-based embedding of a marginalized density, where marginalize out an unspecified object (column) to obtain a marginal density over instances of a specified object (row). Thus, each plot within a single row depicts instances of a single object, but shown differently, dependent on the discovered object. Such a design allows us to analyze how discovered objects impact the objects that are specified by the user.

\textbf{Interactive selection and linked highlighting:} the 1D and 2D density encodings are linked together through interactive brushing.
Specifically, a user can select a subset of scenes for a given object either through brushing one of the violin plots, or brushing one of the scatterplots.
Consequently, we update all views to display the subset, where for scatterplots we highlight the points corresponding to the scenes.
For the violin plots, we first derive object-specific density estimates across all objects, limited to the selected subset. We then encode these subset-restricted densities as new violin plots, and superimpose them on top of their respective objects. Moreover, we populate a scrollable image view with the corresponding scenes, to allow an inspection of the user's selection.
Fig.~\ref{fig:design-b} shows one example of such an interaction.
In selecting a dense cluster over a marginalized density embedding for ``floor'', we find Shih Tzu dogs sitting on pavement; the subset of scenes for the ``paws''-marginalized densities indicates an outlier, highlighting a rare type of paw for a Shih Tzu.

\textbf{Projection steering:} CUPID further allows for a finer-grained analysis of object relationships via the interactive steering of projections.
Specifically, a user may opt to select a set of scenes for a discovered object via the green violin plots, resulting in an update to the image view.
These images serve as a means of finding potential dependencies between objects, via a mechanism for \emph{conditioning}.
Namely, when hovering over an image that contains the object, we show a point corresponding to the detection found through RAM's cross attention.
That specific object, for a given scene, is then used to derive a PMI score for every scene across prompt objects.
We then update the color encoding of the scatterplots -- limited to the discovered object -- by the PMI values. We use a diverging color scale, in order to convey strong co-occurrence between objects (positive value, shown as red), and rare co-occurence (negative value, shown as blue).
If a user clicks on this image, we form a conditional density for each prompt object, conditioning on the selected instance of the discovered object.
We then compute a new 2D density-based embedding, for each of the object prompts, using this conditional embedding as the target density.
Fig.~\ref{fig:design-c} highlights this method of steering, conditioning on a given instance of the ``paw'' object in order to reproject the instances of ``dog'' -- this leads to a grouping of instances that better reflects the conditional density.
\section{Results}

In this section we show the benefits of CUPID in analyzing a variety of prompt-conditioned image distributions.
CUPID's interface is designed to explore a large collection of objects, whether specified or not; but due to limited display, the results presented here correspond to a small set of objects found across images.
We organize our experiments based on the analysis modes discussed in Sec.~\ref{sec:design}, e.g. \emph{verifying} whether the distribution is faithful to the prompt and avoiding biases, as well as a means of \emph{discovering} scenes that exist within a distribution.
Due to space limitations we defer additional results to supplementary material.

\begin{figure}[!t]
   \centering
   \includegraphics[width=.47\textwidth]{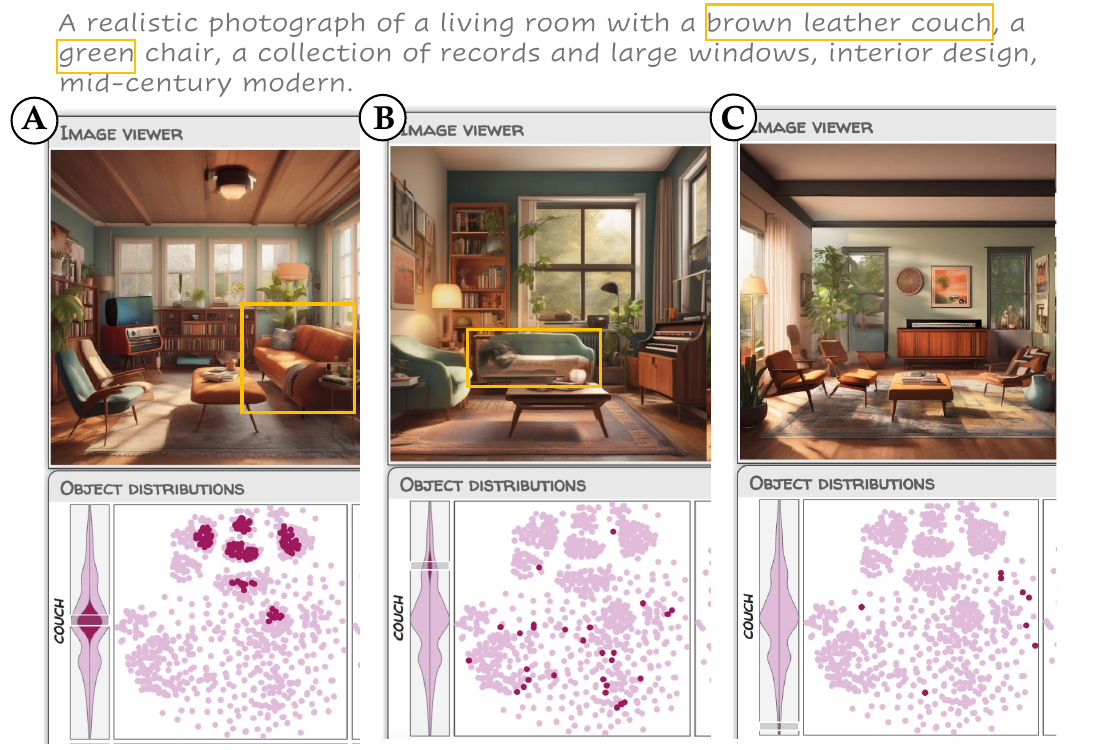}
   \caption{We show a basic use case of CUPID: exploring the density, restricted to a given object. We find that object instances of highest density \textbf{(A)} happen to correspond to objects that are faithfully generated, both in existence (a couch) and appearance (brown leather couch). By selecting a region of smaller density \textbf{(B)}, we find a couch still synthesized, yet its properties do not adhere to the prompt, e.g. green color. Regions of low density \textbf{(C)} can either indicate anomalous objects, or in this case outliers, e.g. no couch present.
   }
   \label{fig:verification_faithfulness}
\end{figure}

\textbf{Experimental details:} for all results in the paper we set the bandwidth in KDE $h = 40$; we find the conclusions made in the results are not sensitive to small variations in the bandwidth setting.
Furthermore, for tSNE, we set the perplexity to $k=7$ for the 1D density embeddings, and $k=14$ for the 2D density embeddings.
Embeddings are optimized starting from a random initialization of positions -- as a result, in comparing 2D projections of a single prompt object across all discovered objects, the scatterplots are not necessarily in alignment.
Nevertheless, we find that linked highlighting helps address the lack of alignment between plots.

We use SDXL~\cite{podell2023sdxl} with its default parameters, and classifier-free guidance scale set to $5$ to strike a balance between prompt adherence and image quality. In all experiments, we draw $1,000$ samples from a prompt-conditioned distribution for visual analysis. For the studied prompts, we find that drawing more samples does not impact the results too much; however in general, the number of samples necessary to ensure sufficient distribution coverage is very likely prompt-dependent. From this sample size, our interface maintains interactivity in (1) linked brushing, and (2) image-conditioned PMI encoding. For projection steering, we find that the optimization process takes around 2 seconds to converge for a single prompt object, thus adding a small amount of latency between projection updates.

\subsection{Verification : faithfulness to objects and their properties}

When exploring images synthesized by a text-to-image model, the most basic questions one might have is: are the generated images consistent with the given prompt? By exploring different modes in the 1D density embeddings of an object in the prompt, users can explore typical and rare images of the object, in order to see if the object is consistently generated by the model, e.g. does the model faithfully synthesize the object's properties based on the prompt?
In Fig.~\ref{fig:verification_faithfulness} we show the ability of CUPID to support this basic kind of analysis.
Specifically, for this description of a living room, we highlight 1D density-based selections for the ``couch'' object. 
We find that for regions where the density is highest, the model synthesizes couches that are consistent with the prompt, e.g. brown and leather.
For less dense regions, we find that a couch is still generated, but its properties are incorrect, e.g. its color is green.
This can be a disadvantage if the user were interested styles that are somewhat rare, relative to the primary mode, yet still consistent with the prompt.
Last, for low-density regions, we find that our density-based method conveys the absence of a couch.

\subsection{Discovery: unspecified objects and properties}

\begin{figure}[!t]
   \centering
       \includegraphics[width=.48\textwidth]{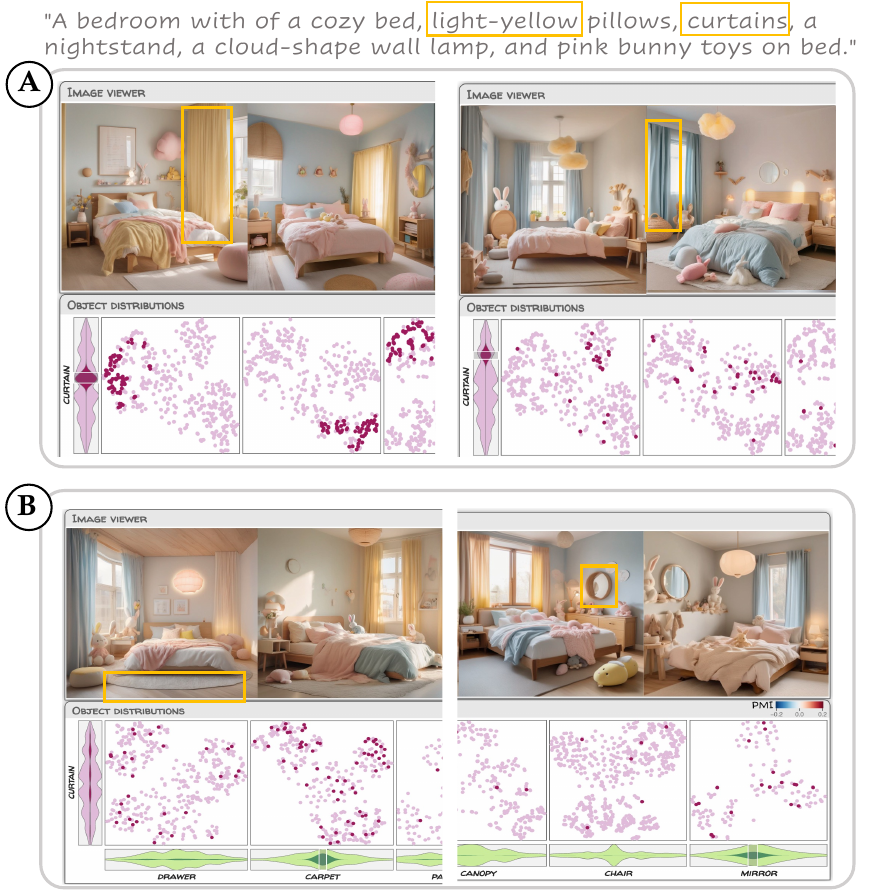}
   \caption{Here we study the relationship between objects, both specified and left unsaid, and their properties, both specified and left unsaid, relative to the prompt (top).
   \textbf{A. }Objects with unspecified properties are assigned with properties in reference to other objects (left) or unspecified properties in the prompt (right)
   \textbf{B. }Unspecified objects are assigned with arbitrary properties not specified in the prompt.
    }
   \label{fig:discover_fig}
\end{figure}

For discovery purposes, one may use CUPID to not merely report objects mentioned, or unspecified, in the prompt, but also find different properties of those objects, using our density-based exploration method.
To demonstrate, in Fig.~\ref{fig:discover_fig} we show an image distribution corresponding to a prompt describing an indoor scene, e.g. a cozy bedroom.
We summarize two types of scenes discovered by CUPID that contain unexpected objects or properties, given the prompt. As shown in scenario (A), the images faithfully contain ``curtains'' which are specified in the prompt, with unstated color.
However, they are assigned to (left) light-yellow color which is made in reference to pillows in the prompt.
This indicates a potential bias of the model, using a specified color in the prompt, for an unspecified object.
On the other hand, on the right we find a blue color is applied, e.g. an arbitrary color not mentioned in the prompt.
In scenario (B), the unspecified objects ``carpet'' and ``mirror'' are both generated with their common characteristics unrelated to the prompt, which introduces more diversity in the synthesized images, without bias.
By exploring these types of images on CUPID, users can have a better understanding the properties that a model decides on, in the absence of the user making these decisions.

\subsection{Discovery: scene composition}

In this section we attempt to discover whether any potential biases might exist in scene composition. We organize this by (1) scene perspective, and (2) spatial relationships.

\begin{figure}[!t]
    \centering
    \includegraphics[width=.42\textwidth]{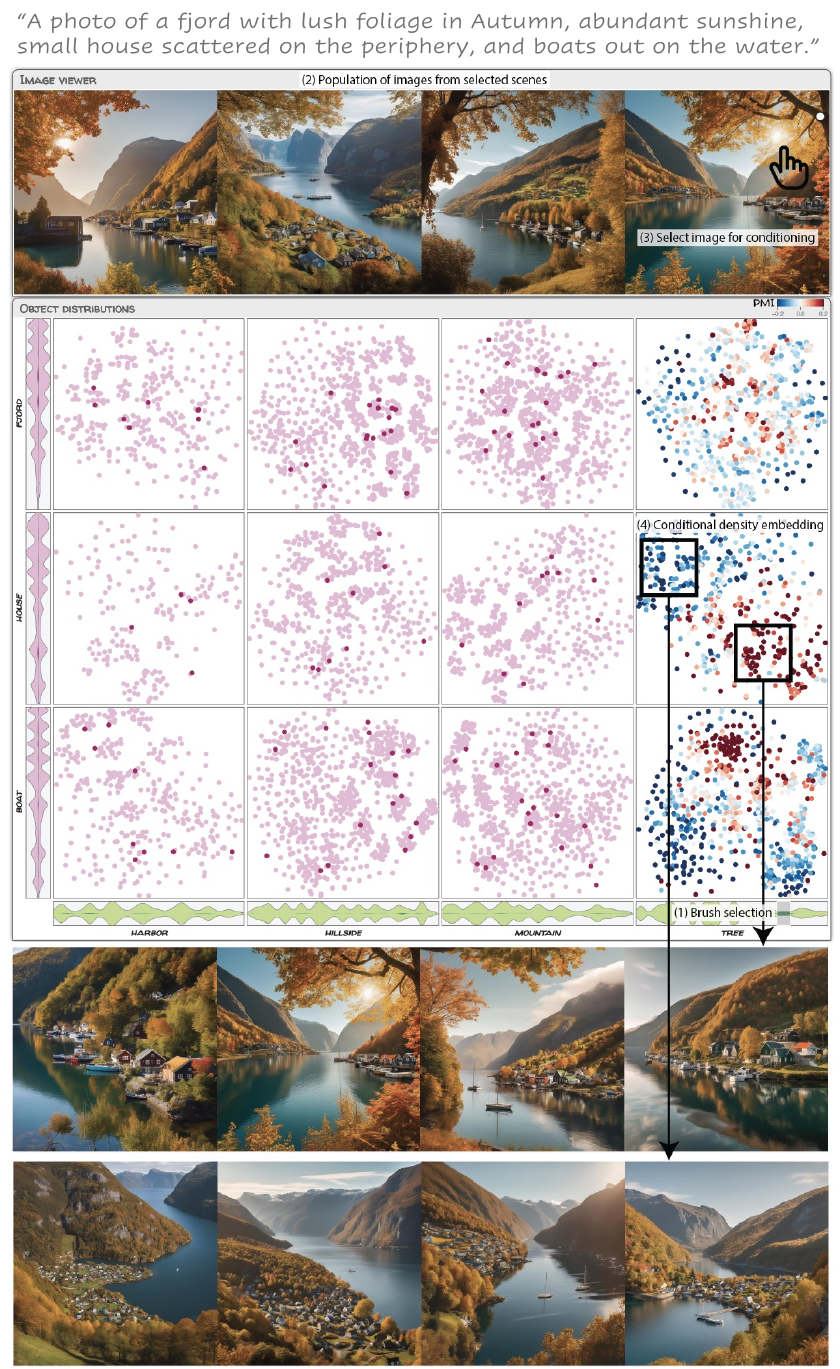}
    \caption{We study object relationships in this distribution of outdoor scenes, wherein upon brushing trees estimated as low density, we find images that contain trees in close view. By selecting an image and showing the PMI between the prompt's objects (right), we find a potential bias in perspective: such trees that are close to the camera are limited to scenes of wide-ranged views.}
    \label{fig:fjord}
\end{figure}

\textbf{Scene perspective:} in Fig.~\ref{fig:fjord} we show an image distribution corresponding to a prompt describing an outdoor scene, effectively, a fjord in autumn.
The model faithfully synthesizes many prompt objects, but also discovers objects not mentioned, e.g. trees, mountains, a harbor.
Here we aim to select \emph{rare types of trees}, shown as a highlighted green selection under the tree's 1D density embedding, where the area mark is indicating a small number of scenes.
We find images corresponding to trees that are in the foreground, e.g. close to an envisioned camera that is taking the supposed photos.

By hovering over one of the images, CUPID then selects the particular tree in this scene to understand potential dependencies in the prompt objects.
Specifically, the PMI scores between this specific tree object, and \emph{all} prompt objects, are computed, and visualized as a color map in the left column.
In turn, the projections are updated based on the conditional distribution, to give a better grouping of objects that have similar dependencies (e.g. grouping objects of similar co-occurrence with the tree object).
In particular, for the house object we see two salient groups of points, which reveal different types of perspective biases in the model.
Specifically, house objects that have negative PMI are shown to be at a far distance from the camera; at the same times, these are scenes for which no trees are close to the camera.
Such an exclusion is not an intrinsic property of the environment, rather, a potential bias learned by the diffusion model based on training data.

\begin{figure}[!t]
    \centering
    \includegraphics[width=.45\textwidth]{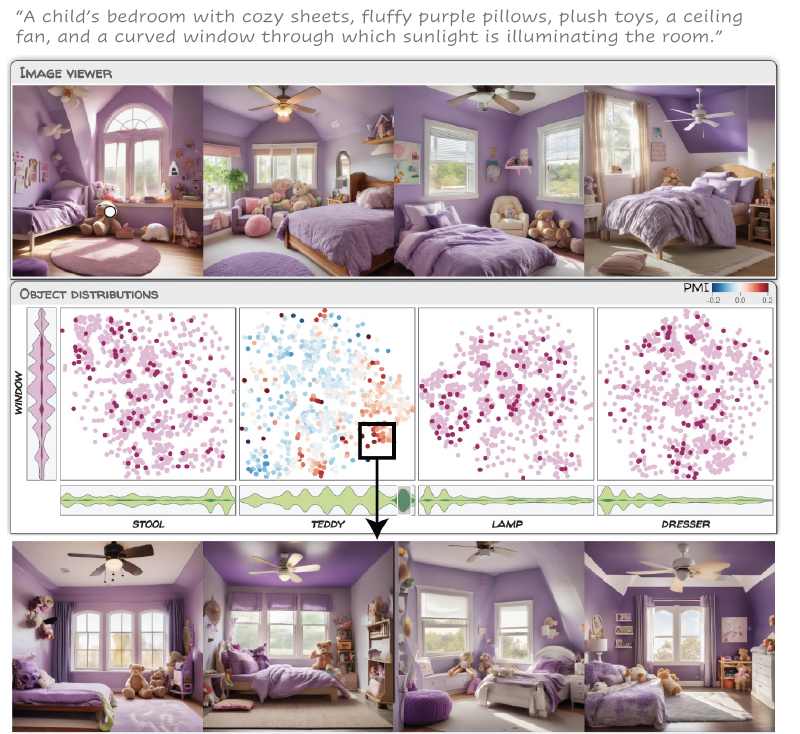}
    \caption{For this distribution of bedroom scenes, we first select a group of objects of teddy bears, finding these correspond to large-sized toys (top). Selecting an image (top left), we see a set of window objects (red color) that contain a strong dependency with the teddy bear. Selecting this group, we find a potential bias in spatial relationships, e.g. teddy bears positioned against or near windows.}
    \label{fig:bedroom}
\end{figure}

\textbf{Spatial relationships:} Fig.~\ref{fig:bedroom} shows an image distribution corresponding to a prompt that describes a child's bedroom.
Through specification of ``toys'' in the prompt, we find a more precise description synthesized by the model: ``teddy'', or teddy bears.
Upon selecting a mode in the distribution for this object, we find that this group of teddys are of large size (top).
Through hovering one of the images, shown in the top-left with the circle indicating the location of detection, we then compute the PMI of all prompt objects, relative to the teddy in this scene.
We find a group of scenes corresponding to windows that indicate a strong co-occurrence, e.g. a PMI value that is positive and large magnitude.
Upon selecting this group, we indeed find teddy bears that tend to be positioned up against, or near, a window.
This indicates a potential bias in spatial relationship, e.g. without explicitly specifying where such big teddy bears should be positioned, we would not necessarily expect so many to appear against windows.
\section{Discussion}

We have shown how our method, CUPID, provides a way to better understand the space of images produced by a text-to-image generative model.
Although we have highlighted the benefits of CUPID, namely the exploration of objects through their densities and discovery of object relationships, we acknowledge several limitations with our method.
The design of CUPID is object-centric: we assume images consisting of objects that can be named, and localized.
As such, CUPID is not designed for arbitrary prompts.
For instance, organizing images along more abstract qualities is not possible, ranging from scene aesthetics, environmental conditions (e.g. weather), and more broadly, image-level rather than object-level features.
We believe such limitations can be addressed by using image-level representations, for instance derived by CLIP~\cite{radford2021learning,goh2021multimodal}, but this requires a different strategy to derive a textual description.
Moreover, the use of RAM~\cite{zhang2023recognize} places limits on what aspects of an object we can represent.
For instance, we cannot easily deploy object representations for a fine-grained understanding of human faces, which would necessitate an adjective (e.g. types of emotion) rather than noun-driven (e.g. anatomical features) approach.
Last, detections found via RAM are imperfect, with the primary issue being false positive detections of objects.
As we demonstrate (c.f. Fig.~\ref{fig:verification_faithfulness}), these often manifest as outliers in low-density regions.
This inhibits us from distinguishing anomalies from outliers, but not at the cost of capturing common objects found in the distribution.

We see a number of promising research directions for future work.
First, we plan to integrate CUPID within existing interfaces for human-AI co-creation~\cite{brade2023promptify,feng2023promptmagician}.
Although the design of the CUPID interface is centered on images drawn from a single prompt-conditioned distribution, our proposed density-based embeddings are a more general approach for deriving visual representations of object-based density.
Such objects can be found in images sampled from multiple distributions, spanning different prompts and model parameters.
Our methods for organizing, and exploring, a distribution can easily extend to more elaborate use cases, e.g. helping a user find a desired image amongst a larger image set, generated under a collection of prompts~\cite{brade2023promptify}.

Additionally, we plan to extend CUPID for model interpretability purposes.
A natural question to consider is the following: what are the similarities/differences between two conditional distributions?
The object-centric approach taken by CUPID lends well to this type of study.
As an example, the distributions might be comprised of the same objects, but the images could vary based on subtle differences in scene descriptions.
We think that purposing CUPID towards comparative visualization designs can help shed light on how language differences manifest as differences between image distributions, and consequently, point towards better guidelines on prompt design.
Moreover, it is often necessary to set model parameters that govern the adherence of generated images to the prompt.
Comparative parameter studies, e.g. comparing two distributions that vary by the strength in guidance scale, can further be investigated via the analysis proposed in CUPID.
More broadly, we believe the methods \& design of CUPID have significant potential for addressing, both, end users in image creation, and model interpretability challenges.
\bibliographystyle{eg-alpha-doi} 
\bibliography{refs}

\end{document}